\documentclass[]{spiejour}  
\usepackage[]{graphicx}

\title{A Dynamical Systems Algorithm for Clustering in Hyperspectral Imagery}

\author{William F. Basener\supscr{a}, Alexey Castrodad\supscr{b}, David Messinger\supscr{c}, Jennifer Mahle\supscr{d} and Paul Prue\supscr{e}}

\affiliation{\supscr{a}Rochester Institute of Technology\\
\linkable{wfbsma@rit.edu} \\
\supscr{b}Department of Defense\\
\supscr{c}Rochester Institute of Technology\\
\linkable{dwmpci@rit.edu} \\
\supscr{d}Florida Atlantic University, Honors College, student\\
\supscr{e}University of California, Davis, student}

\authorinfo{Send correspondence to W. F. B.}

\begin{document}
  \maketitle

\begin{abstract}
In this paper we present a new dynamical systems algorithm for clustering in hyperspectral images. The main idea of the algorithm is that
data points are \`pushed\' in the direction of increasing density and groups of pixels that end up in the same dense regions belong to the
same class. This is essentially a numerical solution of the differential equation defined by the gradient of the density of data points on
the data manifold. The number of classes is automated and the resulting clustering can be extremely accurate. In addition to providing a
accurate clustering, this algorithm presents a new tool for understanding hyperspectral data in high dimensions. We evaluate the algorithm
on the Urban\footnote{Available at \linkable{www.tec.ary.mil/Hypercube/}} scene comparing performance against the $k-$means algorithm using
pre-identified classes of materials as ground truth.
\end{abstract}



\section{INTRODUCTION}\label{sect:intro}
\subsection{Clustering and Classification}
The goal of a clustering algorithm is to put vectors of data into similarity groups.  There are numerous algorithms for clustering
(see~\cite{DJ88} and~\cite{KR90}) the most common of which is the statistical $k$-means algorithm.  In spectral imagery, the goal is to
pixels into groups of like materials. The utility of a clustering algorithm in imagery depends on the goal of the user. For example, an
algorithm that tends towards large clusters may be good for land classification but poor for situations where the user would prefer not to
place isolated man-made objects into large ``background'' classes. If the purpose of clustering is as a pre-processing step for target
detection then the most desired property may be Gaussian distribution of classes while the materials represented in the classes may be
irrelevant. For this paper we assume the primary goal is to cluster pixels to match the various materials represented in an image.

The threshold criteria for grouping two materials in the same cluster is user-dependent.  For example, it might be desirable to have a
single class for trees, or the user might wish to subdivide trees into deciduous and evergreens, or further into individual species. Or,
possibly the user prefers to have a single vegetation class. We will user the term \textit{resolution} to describe the degree to which a
clustering separates materials into different clusters. A clustering subdividing trees into individual species has high resolution whereas
a clustering keeping a single vegetation class has a low resolution. We assume that a user has a predefined degree of resolution; for some
users, deciduous and evergreen trees should be considered as similar, while for other users these should be considered different materials.

Throughout this paper we make a distinction between clusters and classes. A cluster is a group of pixels designated by an algorithm without
regard to the material they represent. A class is a group of pixels that represent a known material. A class may be user-defined, for
example by comparing spectra to a library signature or by hand selecting pixels in the image.

\subsection{Metrics for Evaluating Clusters}
Our goal for clustering can now be stated as follows: A good clustering algorithm should create clusters that closely match known classes
of materials.  It is useful to be more precise when we say clusters ``closely match'' known classes of materials.

A good clustering algorithm should create clusters that are homogeneous (all pixels in the cluster should represent similar materials)
while creating a minimal number of classes for each type of material. We refer to these two characteristics as within \textit{cluster
homogeneity} and material \textit{class preservation}. It should be noted that the degree of class preservation is somewhat dependent on
the classes of materials that we wish to preserve - and thus the desired resolution of the output. The goal is to maximize both cluster
homogeneity and class preservation.

Observe that these two properties are at odds with each other. At one extreme, a clustering putting each pixel in its own cluster will have
high homogeneity (each cluster is very homogeneous) but low preservation (material classes are not preserved - each material class will be
separated over many clusters).  At the other extreme, a clustering putting all pixels in a single cluster will have low homogeneity but
high preservation.

To evaluate our algorithms, we choose ten material classes as ``ground truth" prior to running the algorithms
(Figure~\ref{Fig:class_locations}). This designation of classes can be viewed as determining our desired resolution - we wish to
distinguish between grass and trees, and between various road surfaces, etc. To measure the degree of class preservation of each
clustering, we examine each class and count the number of clusters that the class is split across as well as the percentage of the class
pixels that are placed in each cluster. If there is a cluster that contains all or nearly all pixels from a class then the clustering has
high class preservation for that class. On the other hand, if a class is broken across several clusters than we say that there is low class
preservation.

For each material class, we refer to the cluster containing the most pixels from that class as the \textit{primary cluster} for the class.
The percentage of pixels in the primary cluster for each class is shown in Table~\ref{Table:ClassPreservation}. Observe, for example, that
the $k-$means algorithm with $k=13$ has high preservation of the Wal-Mart roof class but low preservation of the tree class. Note that the
gradient flow algorithm has better class preservation than $k-$means with $k=20$.

To measure homogeneity of our algorithm output, we inspect the primary cluster for each class and determine if this cluster contains a
significant number of pixels from a different class. A cluster that is a primary cluster for one class and also contains a significant
(>5\%) amount of pixels from another class will be called a \textit{mixed cluster}. Table~\ref{Table:ClusterHomogeneity} lists all mixed
clusters along with the percentage of pixels from the non-primary class.

Observe that both the gradient flow algorithm and $k-$means (with $k=13$) mix the Wal-Mart class and gray roof classes as well as the two
road classes. The $k-$means algorithm has more mixing between the roads and roof classes, and thus lower homogeneity.

To avoid hiding the warts and problems of an algorithm behind our choice of metrics, we include a more detailed study of the classes in
Section~\ref{Sec:DetatiledInspection}.  The reader can determine if these metrics accurately represent the performance of the two
algorithms.

From our point of view, the resolution of a clustering - the degree to which the clustering distinguished between various materials - is a
natural parameter to have the user set in an unsupervised clustering algorithm. In $k-$means clustering this is set by determining the
number of classes. In our algorithm, the resolution is determined by the number of neighbors used to construct the graph and the amount of
smoothing. The number of neighbors determines the minimal size of a cluster and the number of smoothing steps determines the tendency to
pull groups of pixels into large clusters. The algorithm is a partial differential equation (PDE) method, in which case these two
parameters correspond to the rate of diffusion and length of time that we run our PDE. The choice of parameters (e.g. 38 smoothing steps in
the gradient flow algorithm and $k=20$ in k-means) were chosen to get a good level of class preservation on the ``ground truth" material
classes. This methodology - choosing parameters to get a desired level of class preservation on a supervised training set - may be a
suitable methodology for semi-automating parameter choices in clustering algorithms in general.

Graph theory has proven a useful tool for analysis of hyperspectral imagery.  The typical method is to construct a graph to model the
hyperspectral data with a vertex at each data point and edges connecting nearby vertices.  These algorithms do not assume that the data is
Gaussian, linear, or even convex.  This enables increased accuracy over traditional methods, at a potential cost of robustness (not having
a model to fit the data to may require more free parameters) and computational time.

Examples that use the graph to approximate a manifold structure include the local linear embedding, Laplacian Eigenmaps, diffusion
Eigenmaps, and ISOMAP. ISOMAP is used effectively in~\cite{BAF05} for dimensionality reduction and to determine geodesic distances between
points for clustering and land classification. In this context, the geodesic distance between two pixels and is the length of the shortest
path from to where the path is restricted to travel along the edges of the graph. The Local Linear Embedding algorithm~\cite{SR03} is
useful for dimension reduction subsequent classification.  A similar approach is used in~\cite{CCG05} for land cover classification.  The
methods can be combined with spatial information for improved performance as in~\cite{MSB07}.  A large comprehensive review of manifold
learning and dimension reduction algorithms is given in~\cite{MPH07}.  The graph theory model has been used effectively as a background
model without assuming the data lie on a manifold in~\cite{BIM07} and~\cite{BM09}.


\section{Gradient Flow Clustering}\label{Sec:GradientFlow}
The motivation for the algorithm, sliding points through the data in the direction of increasing density, is simple and summarized in
Figure~\ref{Fig:PictoralGradientFlow}. The details of the algorithm are discussed in this section and the results will be presented in
Section~\ref{Sec:Results}.

\subsection{Gradient Differential Equations}
For motivation, consider a real-valued function $f$ defined on the Euclidean plane.  The gradient differential equation of $f$ is defined
to be
\[
\frac{\textrm{d} x}{\textrm{d} t} = -\nabla f.
\]
If we think of $f$ as the height of a surface above the plane, then the vector $-\nabla f$ points in the direction of steepest descent. The
maxima of $f$ are called \textit{sources} for the gradient differential equation because the direction of movement is outward from the
maxima, and the minima of $f$ are called \textit{sinks}.  If were to place water on the surface, it would run downhill away from the
sources and into the sinks. For each sink $x$, the set of all initial conditions $y$ such that the solution beginning at $y$ ends at $x$ is
called the \textit{basin of attraction} for $x$.  Following the flowing water analogy, the basin of attraction of a sink $x$ is the place
from which water will run down into $x$.

\begin{figure}
   \begin{center}
   \begin{tabular}{c}
   \includegraphics[height=3in]{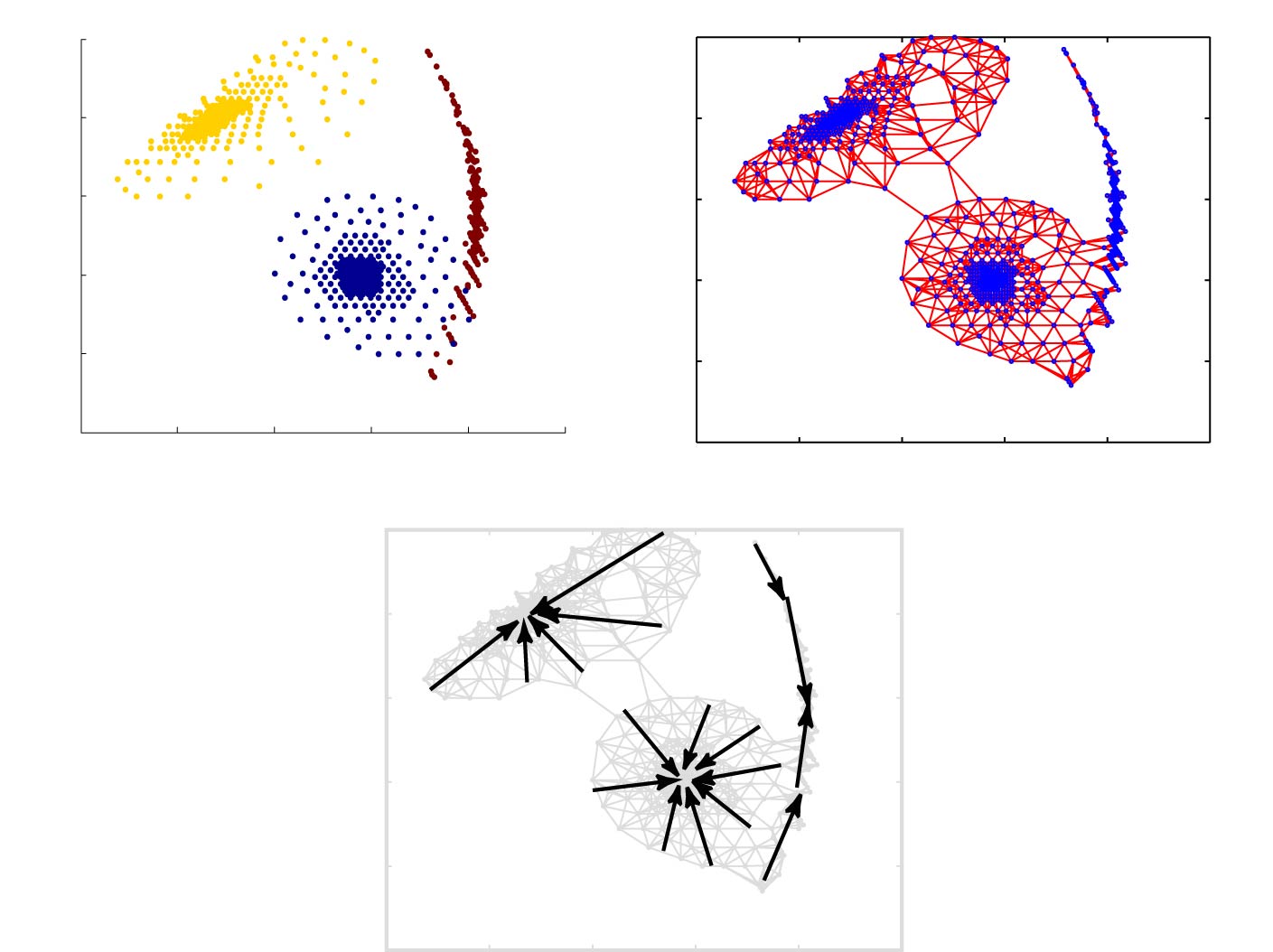}
   \end{tabular}
   \end{center}
   \caption[example]
   { \label{Fig:PictoralGradientFlow}
(Upper left) Three synthetically generated classes. (Upper right) The graph generated by connecting each point to its 6 nearest neighbors.
(Lower) The arrows indicate the direction of movement in the gradient flow. }
\end{figure}

Our algorithm is simple to state using gradient flows.  For a data set in $n-$dimensional space, we define a function that corresponds to
the density of data points around each data point. Then we define the discrete version of the gradient flow differential equation and the
sinks are the maxima for the density function and the clusters are the basins of attraction for the sinks.  See
Figure~\ref{Fig:PictoralGradientFlow} for a pictorial representation of this in $2-$dimensions.

\subsection{The Gradient Flow Algorithm}
In this section we describe the implementation of the gradient flow on an HSI data set.

\noindent\textsc{Step 1: Define the Nearest Neighbor Graph.} For every point $x$ in the data set, compute the $k$ (results shown in this
paper used $k = 40$) nearest neighbors in spectral space. The result is an $n \times k$ matrix $KN$ where $n$ is the number of pixels in
the image. The $i, j^\textrm{th}$ entry in $KN$ is the index for point $i$'s $j^\textrm{th}$ nearest neighbor. This computation provides
the edges in our graph shown in Figure~\ref{Fig:PictoralGradientFlow} as well as the lengths of the edges.

\noindent\textsc{Step 2: Compute the Density Function.}  If $D$ is the matrix whose $i,j^\textrm{th}$ term is the distance from point $i$
to $NK(i,j)$, we take the termwise exponential of $-D^2$ (termwise squaring of $D$),
\[
K = e^{-D\times D/\sigma^2},
\]
where $\sigma$ is the mean value in $D$.  Observe that $K(i,j) = e^{-D(i,j)*D(i,j)/\sigma^2} \in (0,1]$, so $K(i,j)$ can be thought of as a
normalized measure of distance with $i,j$ close corresponding to $K\approx 1$ and $i,j$ far apart corresponding to $K\approx 0$. Define $S$
to be the $n\times 1$ vector obtained by summing $K$ along its rows. Then $S(i)$ gives a measure of the density of data near point $i$:
$S(i)\in (0,k]$ with higher values corresponding to more dense data.

Typically, the $k^\textrm{th}-$\textit{codensity} of a point in a data set is defined to be the distance to its $k^\textrm{th}$ nearest
neighbor, and hence is the reciprocal of the density; codensity is the radius of a ball containing $k$ neighbors whereas density is the
number of neighbors in a ball of radius $r$. Our vector $S$ gives a measure of density that takes into account the distance to neighboring
points as well as the number of neighboring points.

\noindent\textsc{Comment:} The \textsc{Matlab} module DSTool, which is available for public download, has an implementation of an algorithm
that computes $KN$ and $D$ relatively quickly.  Once $KN$ and $D$ are computed, the remained of the steps in the algorithm are fast with a
total runtime of less than a few seconds.

\noindent\textsc{Step 3: Smoothing.} We wish to identify not just points of high density, but points in centers of regions of high density.
To measure this trait we define a new density function $SS$ (Sum of S) such that $SS(i)$ is the sum of $S$ at the $k-$nearest
neighbors of point $i$, which is the sum across the rows of $S(KN)$. We can iterate this summing any number of times. More iterations
corresponds to more smoothing, or in terms of PDEs, running the PDE for more time.

\noindent\textsc{Step 4:  The Gradient ODE.} We define a map $F$ from the data set to itself such that $F(x)$ is equal to the neighbor of
$x$ with the smallest value of $SS$. (In the code, $F(i)$ is the index $KN(i, j)$ for which $SS(KN(i, j))$ is minimal.) Then the map $F$,
represented as an $n-$dimensional vector, is the discretization of the gradient differential equation. To iterate the differential
equation, we simply run $F = F(F)$, and repeat until the values of $F$ reach a steady state. This typically takes $5-10$ iterations and
less than a second of runtime.

\noindent\textsc{Step 5: Define the Classes.} After the steady state is reached, the value of $F(i)$ is the index for the minima of $SS$
that the initial condition $i$ reaches. Thus, $F$ is the class map: $F(i) = F(j)$ if and only if point $i$ and point $j$ are in the same
class. The last step is to reform $F$ into an image and choose a colormap to display the classes.


\section{Results}\label{Sec:Results}

\subsection{Overview} 
\begin{figure}[h]
   \begin{center}
   \begin{tabular}{c}
   \includegraphics[height=2in]{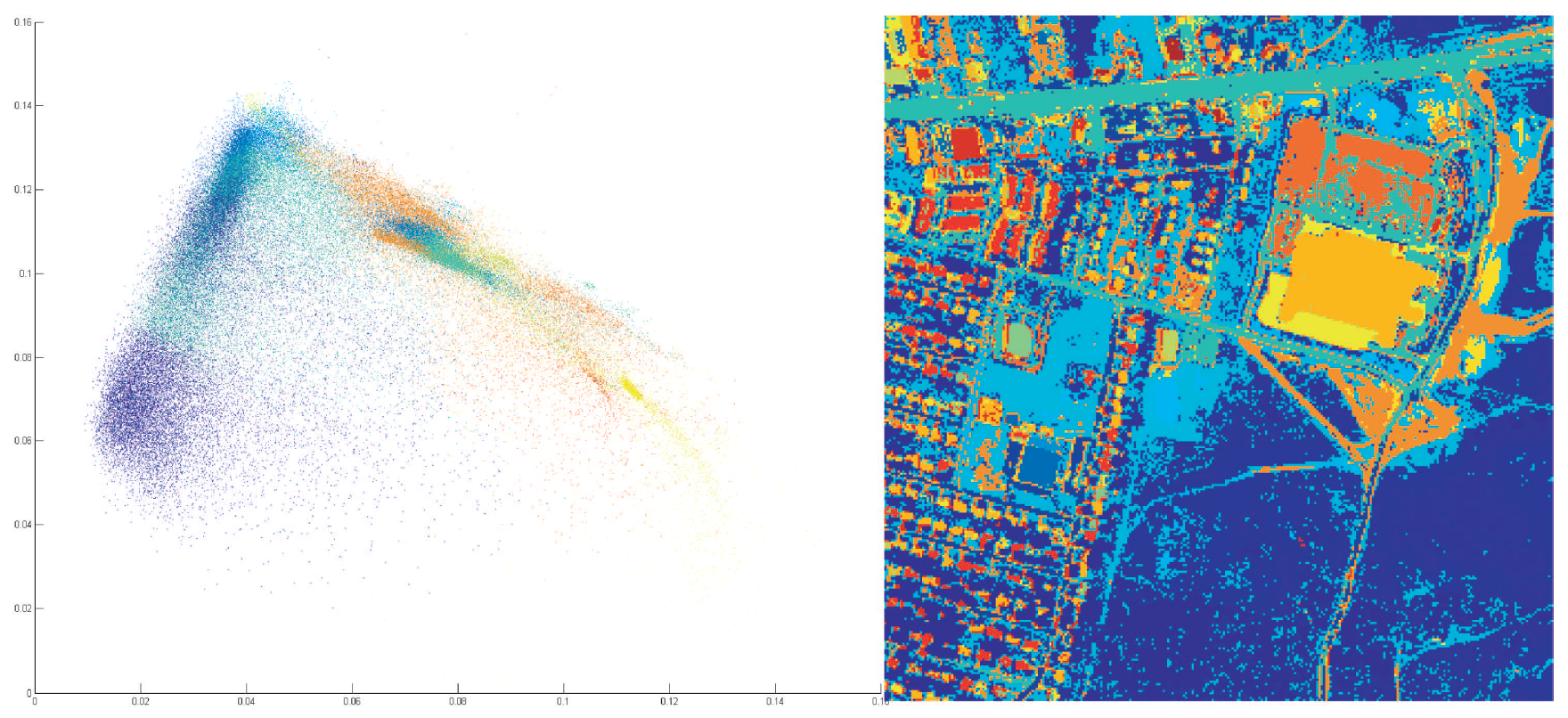}
   \end{tabular}
   \end{center}
   \caption[example]
   { \label{Fig:GradientFlowClusters}
(Left) The scatter plot showing the classes from the gradient flow algorithm. (Right) A class map showing the classes on the image. (NOTE:
The colors do not distinguish all major classes.)}
\end{figure}

\begin{figure}[h]
   \begin{center}
   \begin{tabular}{c}
   \includegraphics[height=2in]{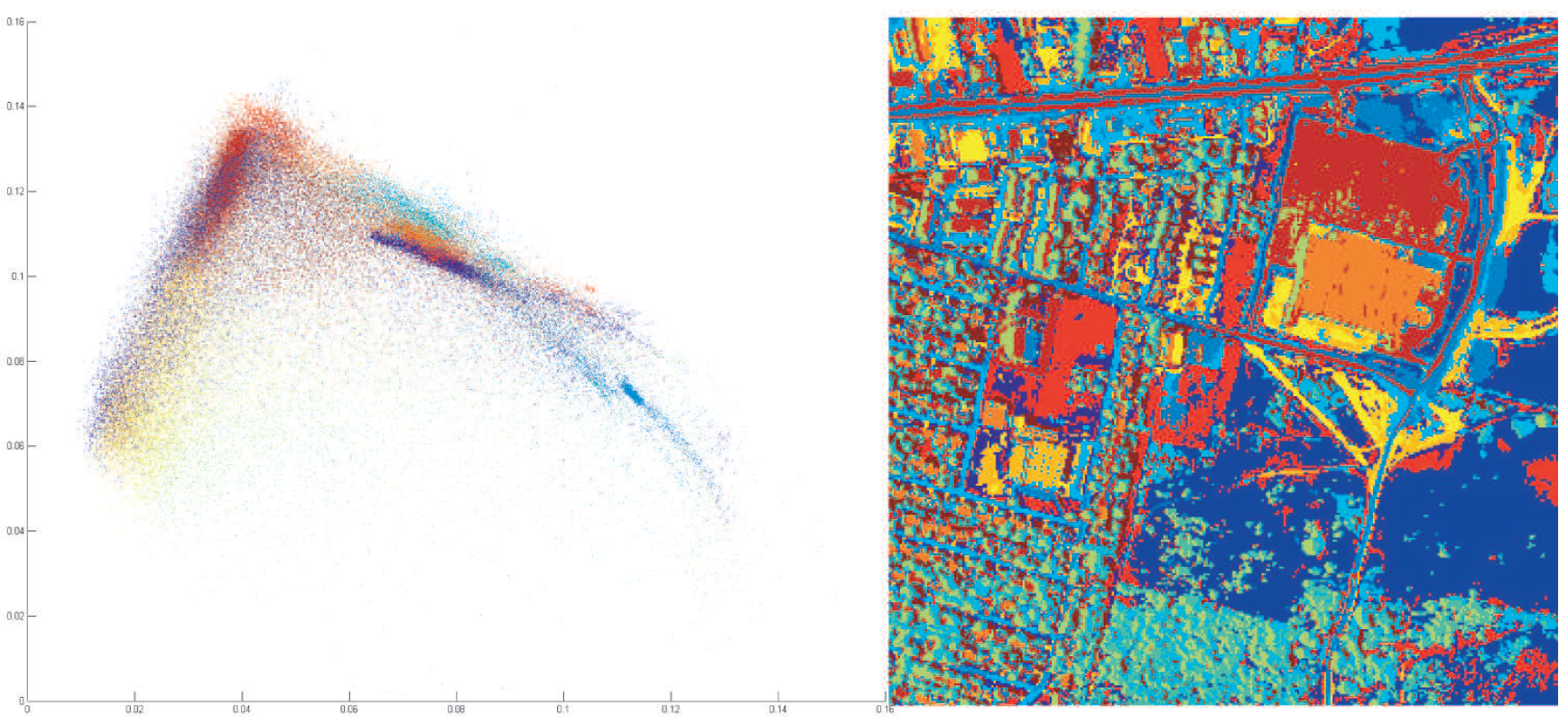}
   \end{tabular}
   \end{center}
   \caption[example]
   { \label{Fig:K-MeansClusters}
(Left) The scatter plot showing the classes from the $k-$means algorithm. (Right) A class map showing the classes on the image. (NOTE: The
colors do not distinguish all classes.)}
\end{figure}

We evaluate our algorithm on the Urban image, which is 300 samples by 307 lines with 117 bands. The result from running the
gradient flow algorithm and $k-$means ($k = 20$) are shown in Figures~\ref{Fig:GradientFlowClusters} and~\ref{Fig:K-MeansClusters}.

The standard metric for comparing a clustering to a ground truth classification mask is to designate a single cluster associated with each
ground truth class and compute the percentage of all pixels in the regions covered by the ground truth classes that are designated into the
correct clusters, called the percent accuracy.  Our ground truth is shown in Figure~\ref{Fig:class_locations} and the percent accuracy for
gradient flow is 85.9\% and the percent accuracy for $k-$ means with 20 classes is 71.9\%.

One could argue that the usual percent accuracy metric is easy to construe: a clustering with too high or too low resolution will score
poorly. A more detailed comparison between clustering and ground truth is given by our class-preservation and cluster homogeneity metrics.
Recall that a high resolution (many small clusters) will have low class-preservation and high cluster homogeneity while a low resolution
clustering (few, large clusters) will have high class preservation and low cluster homogeneity.  The goal is to cluster an image with
simultaneous high class preservation and cluster homogeneity.  The primary conclusion of this section is that the gradient flow algorithm
with 20 clusters has better class-preservation (see Table~\ref{Table:ClassPreservation}) and better cluster homogeneity (see
Table~\ref{Table:ClusterHomogeneity} than $k-$means with 20 clusters.

\subsection{Comparison to Ground Truth Classes}
\begin{figure}[h]
   \begin{center}
   \begin{tabular}{c}
   \includegraphics[height=2in]{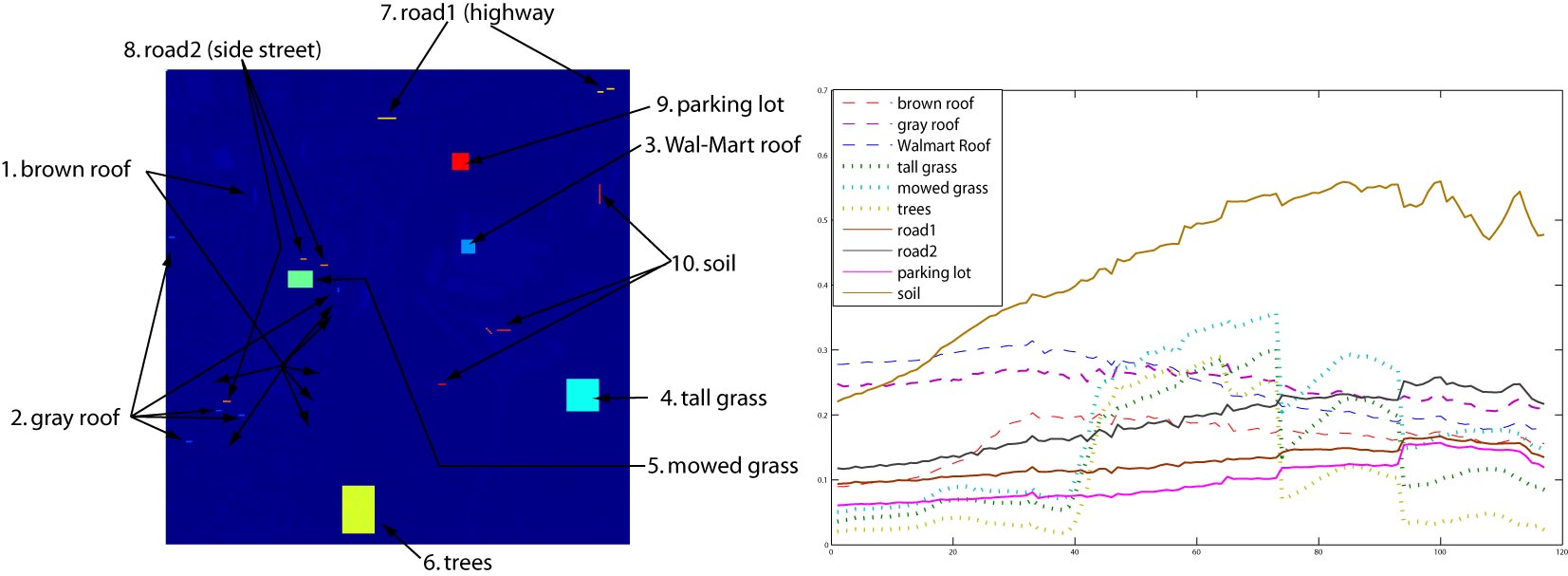}
   \end{tabular}
   \end{center}
   \caption[example]
   { \label{Fig:class_locations}
(left) The locations of the selected pixels for the test classes. (right) The mean spectra for each class.}
\end{figure}

Figure~\ref{Fig:class_locations} shows ten different classes of materials selected in the scene by visual inspection.  The locations of the
pixels for each class are shown along with the mean spectra. Each spectrum was visually inspected to verify that the selected pixels in
each class represent very similar materials.

An attempt was made to keep each class homogeneous but not so homogenous that the classes should be trivial to distinguish. The tall grass
was from what appears to be a region in the image with low variation, each each road class was taken on multiple roads of apparently
similar type, roof classes 1 and 2 include pixels from roofs of multiple houses, and the Wal-Mart roof class did not include pixels on roof
vents and other similar objects. Thus a good clustering algorithm should put most pixels in each class into the same cluster. The potential
exceptions are the roof classes.  Observe, in particular, that the mean spectra for the Wal-Mart roof class appears similar to the mean
spectra for the gray roof class.

\subsection{Material Class Preservation.}\label{SSEC:Material_Class_Preservation}

\begin{figure}[h]
   \begin{center}
   \begin{tabular}{c}
   \includegraphics[width=5in]{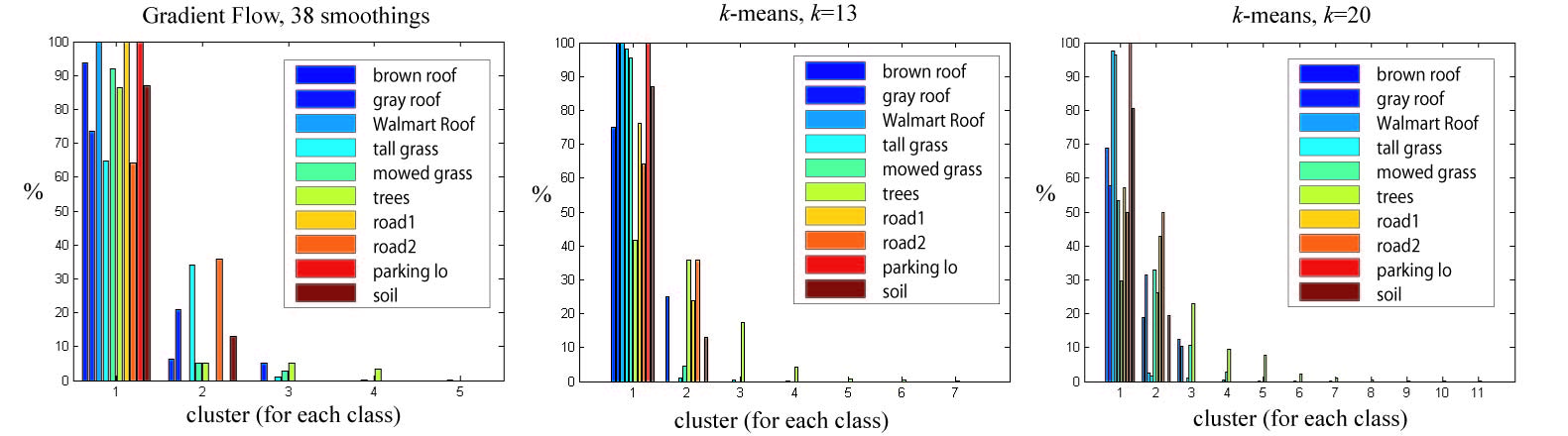}
   \end{tabular}
   \end{center}
   \caption[example]
   { \label{Fig:class_breakdown}
Each bar graph shows the distribution of the ten classes across the clusters containing pixels from that class.}
\end{figure}

One metric for evaluation of a clustering algorithm is the tendency to put the majority of the pixels from each class in a single cluster,
which we call class-preservation.  For each class of pixels, we define the \textit{primary cluster} for that class to be the cluster
containing the largest number of pixels from the class.  We quantitatively test the algorithms with respect to this metric by computing,
for each class, the percentage of pixels from that class that are placed in its primary cluster.  Table~\ref{Table:ClassPreservation} shows
the percentage of pixels from each class that belong to its primary cluster, so higher percentages indicate higher class-preservation. The
full detailed breakdown of each class across all clusters is given in Figure~\ref{Fig:class_breakdown}. Note that the gradient flow
algorithm divided a test class across at most five clusters while $k-$means divided the tree class over 11 clusters.

\begin{table}[h] \caption{Percentage of pixels from each class that are placed in the largest cluster. Higher percentages correspond to better class preservation.} \label{Table:ClassPreservation}
\begin{center}
\begin{tabular}{|l|c c|} 
\hline
\rule[-1ex]{0pt}{3.5ex}   & \multicolumn{2}{|c|}{\textbf{\% of pixels in largest cluster}} \\
\vspace{-4pt}
\rule[-1ex]{0pt}{3.5ex}   & Gradient Flow  & $k-$means \\
\rule[-1ex]{0pt}{3.5ex} \textbf{Class}   & &   ($k=20$) \\
\hline\hline
\rule[-1ex]{0pt}{3.5ex}  1. Brown Roof & 94 \%  &   69 \% \\
\rule[-1ex]{0pt}{3.5ex}  2. Gray Roof & 74 \%  &   58 \% \\
\rule[-1ex]{0pt}{3.5ex}  3. Wal-Mart Roof & 100 \%  &   98 \% \\
\rule[-1ex]{0pt}{3.5ex}  4. Tall Grass & 65 \%  &   96 \% \\
\rule[-1ex]{0pt}{3.5ex}  5. Mowed Grass & 92 \%  &   53 \% \\
\rule[-1ex]{0pt}{3.5ex}  6. Trees & 86 \%  &   30 \% \\
\rule[-1ex]{0pt}{3.5ex}  7. Road 1 (highway) & 100 \%  &   57 \% \\
\rule[-1ex]{0pt}{3.5ex}  8. Road 2 (side Street) & 64 \%  &   50 \% \\
\rule[-1ex]{0pt}{3.5ex}  9. Parking Lot & 100 \%  &   100 \% \\
\rule[-1ex]{0pt}{3.5ex}  10. Soil & 87 \%  &   81 \% \\
\hline
\rule[-1ex]{0pt}{3.5ex}  \textbf{Mean} & 86.2 \%  &   69.1 \% \\
\hline
\end{tabular}
\end{center}
\end{table}

\subsection{Homogeneity.}\label{SSEC:Homogeneity} The second metric we use is the tendency to create clusters containing pixels from more than one
class.

Both algorithms placed the gray Wal-Mart roof in the same class with the gray roofs from houses.  This is reasonable given the similarity
between the gray roof and Wal-Mart roof spectra in Figure~\ref{Fig:class_locations}.  Both algorithms also had difficulty distinguishing
between the two road surfaces (side streets and highways). There was no other significant misclassification by the Gradient Flow
algorithms.  The $k-$means algorithm had additional clusters that contained significant numbers of pixels from various classes. For
example, primary brown roof class for the $k-$means algorithm contains pixels on streets (See Figure~\ref{Fig:1brownroof_kmeanseps}.) In
fact, this cluster is the primary cluster for both the brown roof class and the road 2 class.

We evaluate the algorithms quantitatively as follows. A cluster that is a primary cluster for one class and also contains a large number of
pixels from another class will be called a mixed cluster. We list the mixed clusters (primary cluster for one class containing greater than
5\% of the pixels from a different class) in Table~\ref{Table:ClusterHomogeneity}, giving also the percentage of pixels from the
non-primary class.
\begin{table}[h]
\caption{Mixed clusters - primary clusters for one class that also contain a significant (greater than 5\%) number of pixels from another
class.} \label{Table:ClusterHomogeneity}
\begin{center}
\begin{tabular}{|l|c|c|} 
\hline
\rule[-1ex]{0pt}{3.5ex}  \textbf{Class}  & \textbf{Gradient Flow Mixed Clusters} &  \textbf{$k-$Means Mixed Clusters} \\
\hline\hline
\rule[-1ex]{0pt}{3.5ex}  1. Brown Roof & $<5\%$  &  Road 1 (23 \% ) \\
\rule[-1ex]{0pt}{3.5ex}    &   &  Road 2 (64 \% ) \\
\rule[-1ex]{0pt}{3.5ex}  2. Gray Roof & Wal-Mart (100\% )  &  Wal-Mart (100 \% ) \\
\rule[-1ex]{0pt}{3.5ex}  3. Wal-Mart Roof & Gray Roof (74 \% )  &  Gray Roof (100 \% ) \\
\rule[-1ex]{0pt}{3.5ex}  4. Tall Grass & $<5\%$  &  $<5\%$  \\
\rule[-1ex]{0pt}{3.5ex}  5. Mowed Grass & $<5\%$  &  $<5\%$ \\
\rule[-1ex]{0pt}{3.5ex}  6. Trees & $<5\%$ & $<5\%$ \\
\rule[-1ex]{0pt}{3.5ex}  7. Road 1 (highway) & Road 2 (64\% )  &  Brown Roof (25 \% ) \\
\rule[-1ex]{0pt}{3.5ex}   &  &  Parking Lot (100 \% ) \\
\rule[-1ex]{0pt}{3.5ex}  8. Road 2 (side Street) & Road 1 (100 \% )  &  Road 1 (36 \% ) \\
\rule[-1ex]{0pt}{3.5ex}  9. Parking Lot & $<5\%$  &  Road 1 (76 \% ) \\
\rule[-1ex]{0pt}{3.5ex}  &  &  Brown Roof (25 \% ) \\
\rule[-1ex]{0pt}{3.5ex}  10. Soil & $<5\%$  &  $<5\%$ \\
\hline
\end{tabular}
\end{center}
\end{table}

Observe that the mixing in the gradient flow clusters is between similar materials; Wal-mart Roof is similar to other Gray Roofs, and Road1
is relatively similar to Road 2. The $k-$means algorithm has mixing between diverse materials; most significantly between parking lots,
roofs, and roads.

\section{Detailed inspection of clusters.}\label{Sec:DetatiledInspection}
We qualitatively evaluate performance with respect to this metric by inspecting the clusters based on spatial distribution in the scene and
spectra. For each class we show the two clusters containing most of the pixels from that class. This is shown in
Figures~\ref{Fig:1brownroof_kmeanseps} through~\ref{Fig:10soil_gradientflow}.

\newpage
\subsection{Brown Roof}
\begin{figure}[h]
   \begin{center}
   \begin{tabular}{c}
   \includegraphics[width=3.7in]{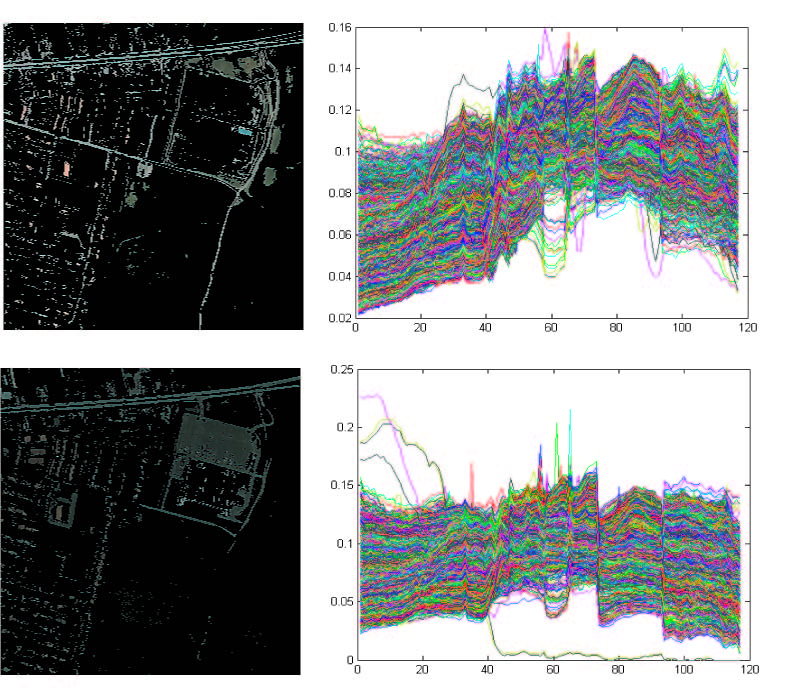}
   \end{tabular}
   \end{center}
\vspace{-10pt}
   \caption{ \label{Fig:1brownroof_kmeanseps}
$k$-means clusters containing pixels from the Brown Roof Class.}
\end{figure}

\begin{figure}[h]
   \begin{center}
   \begin{tabular}{c}
   \includegraphics[width=3.7in]{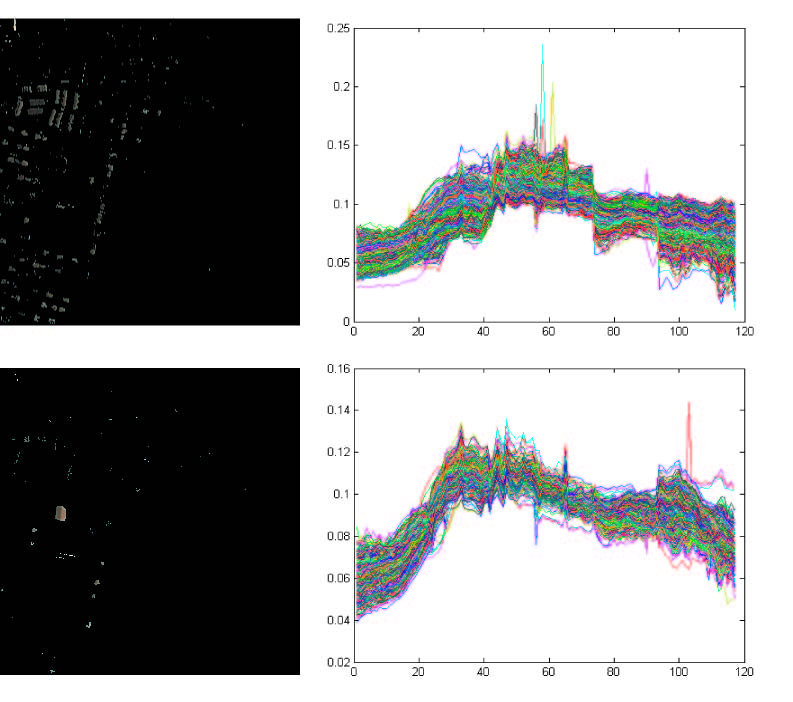}
   \end{tabular}
   \end{center}
\vspace{-10pt}
   \caption{ \label{Fig:1brownroof_gradientflow}
Gradient Flow clusters containing pixels from the Brown Roof Class.}
\end{figure}

Figures~\ref{Fig:1brownroof_kmeanseps} and~\ref{Fig:1brownroof_gradientflow} show the clusters containing pixels from the Brown Roof Class.
Observe that the $k-$means clusters are a mix with roofs and road, and the spectra and pixel locations indicate that the gradient flow
clusters are more homogeneous.
\vspace{-10pt}

\newpage
\subsection{Gray Roof}
\begin{figure}[h]
   \begin{center}
   \begin{tabular}{c}
   \includegraphics[width=3.7in]{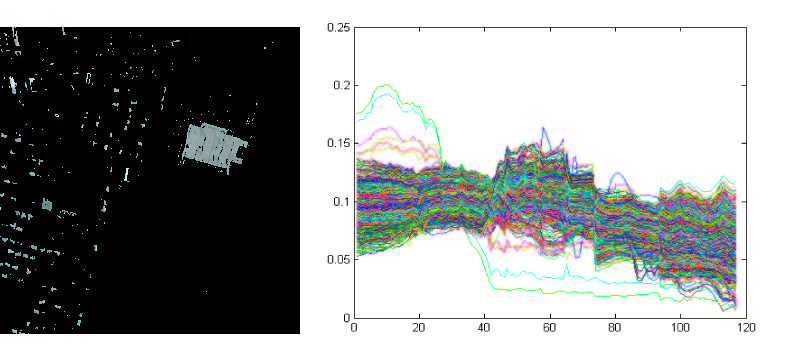}
   \end{tabular}
   \end{center}
\vspace{-10pt}
   \caption{ \label{Fig:2grayroof_kmeans}
$k$-means cluster containing pixels from the Gray Roof Class.}
\end{figure}

\begin{figure}[h]
   \begin{center}
   \begin{tabular}{c}
   \includegraphics[width=3.7in]{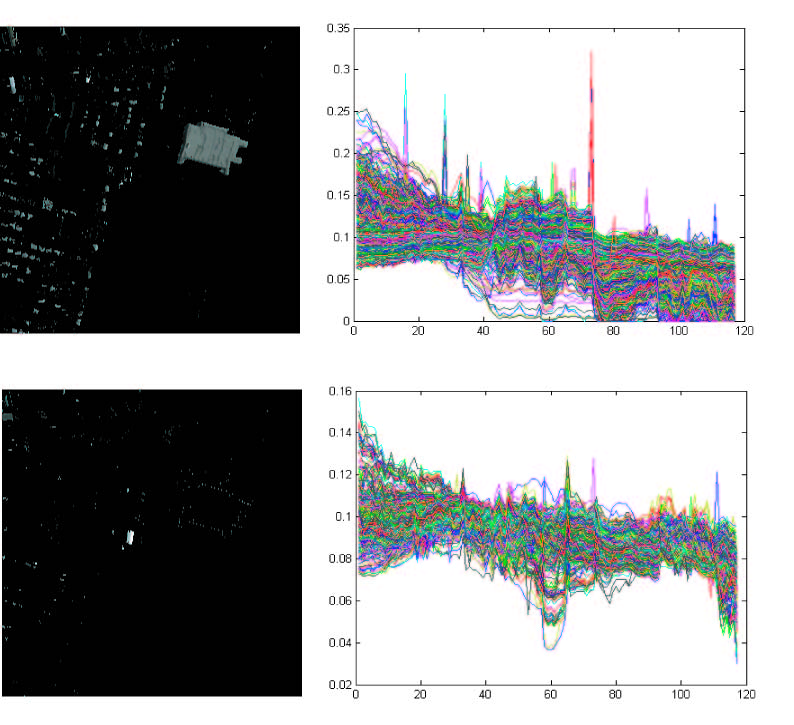}
   \end{tabular}
   \end{center}
\vspace{-10pt}
   \caption{ \label{Fig:2grayroof_gradientflow}
Gradient Flow clusters containing pixels from the Gray Roof Class.}
\end{figure}

Figures~\ref{Fig:2grayroof_kmeans} and~\ref{Fig:2grayroof_gradientflow} show the clusters containing pixels from the Gray Roof Class.  Both
algorithms perform reasonably.
\vspace{-10pt}

\subsection{Wal-Mart Roof}
All of the Wal-Mart Roof pixels are clustered with the gray roof pixels.  There appears to be little spectral difference between the gray
Wal-Mart roof pixels and other gray roof pixels.

\newpage
\subsection{Tall Grass}
\begin{figure}[h]
   \begin{center}
   \begin{tabular}{c}
   \includegraphics[width=3.7in]{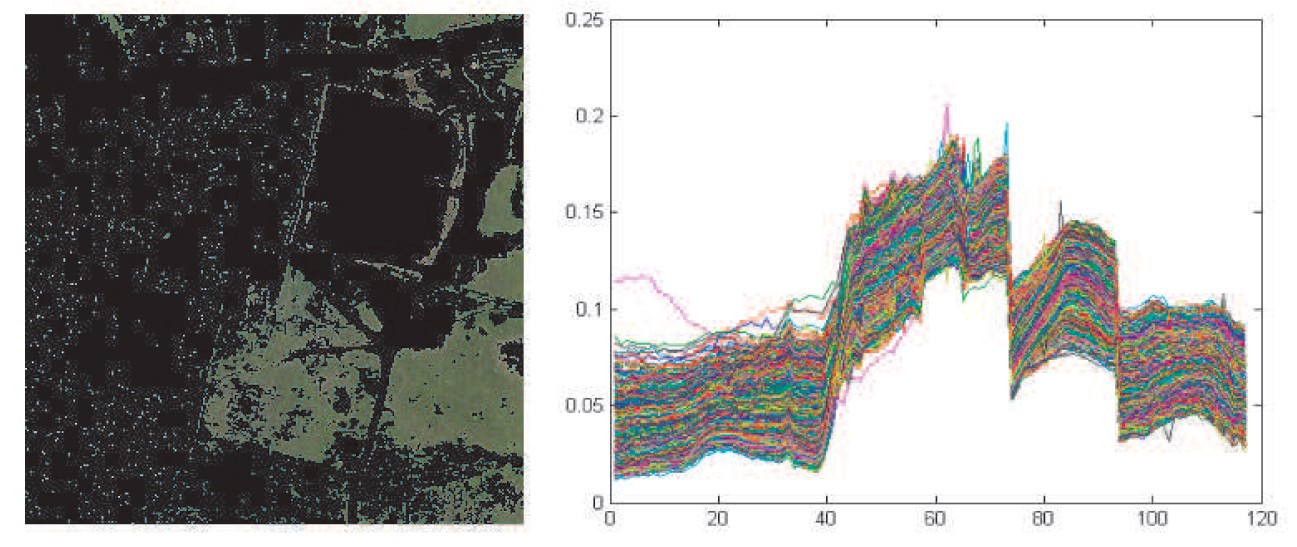}
   \end{tabular}
   \end{center}
\vspace{-10pt}
   \caption{ \label{Fig:4tallgrass_kmeans}
$k$-means cluster containing pixels from the Tall Grass Class.}
\end{figure}

\begin{figure}[h]
   \begin{center}
   \begin{tabular}{c}
   \includegraphics[width=3.7in]{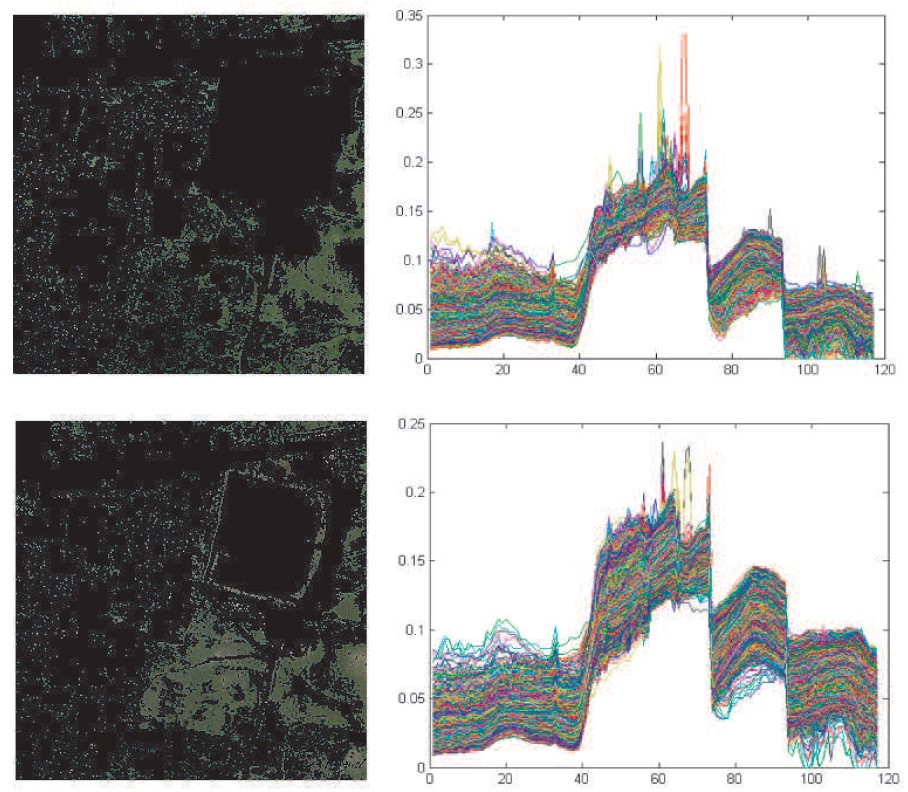}
   \end{tabular}
   \end{center}
\vspace{-10pt}
   \caption{ \label{Fig:4tallgrass_gradientflow}
Gradient Flow clusters containing pixels from the Tall Grass Class.}
\end{figure}

Figures~\ref{Fig:4tallgrass_kmeans} and~\ref{Fig:4tallgrass_gradientflow} show the clusters containing pixels from the Tall Grass Class.
Both algorithms perform reasonably.
\vspace{-10pt}

\newpage
\subsection{Mowed Grass}
\begin{figure}[h]
   \begin{center}
   \begin{tabular}{c}
   \includegraphics[width=3.7in]{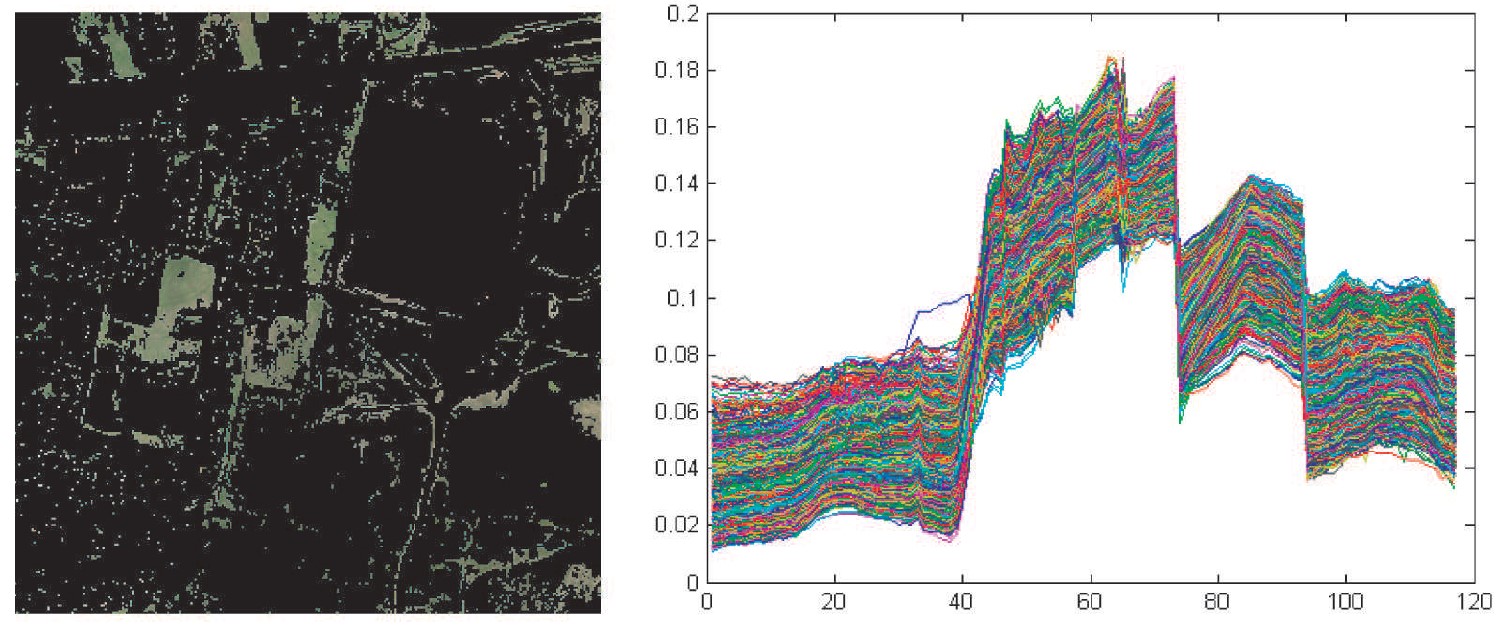}
   \end{tabular}
   \end{center}
\vspace{-10pt}
   \caption{ \label{Fig:5mowedgrass_kmeans}
$k$-means cluster containing pixels from the Mowed Grass Class.}
\end{figure}

\begin{figure}[h]
   \begin{center}
   \begin{tabular}{c}
   \includegraphics[width=3.7in]{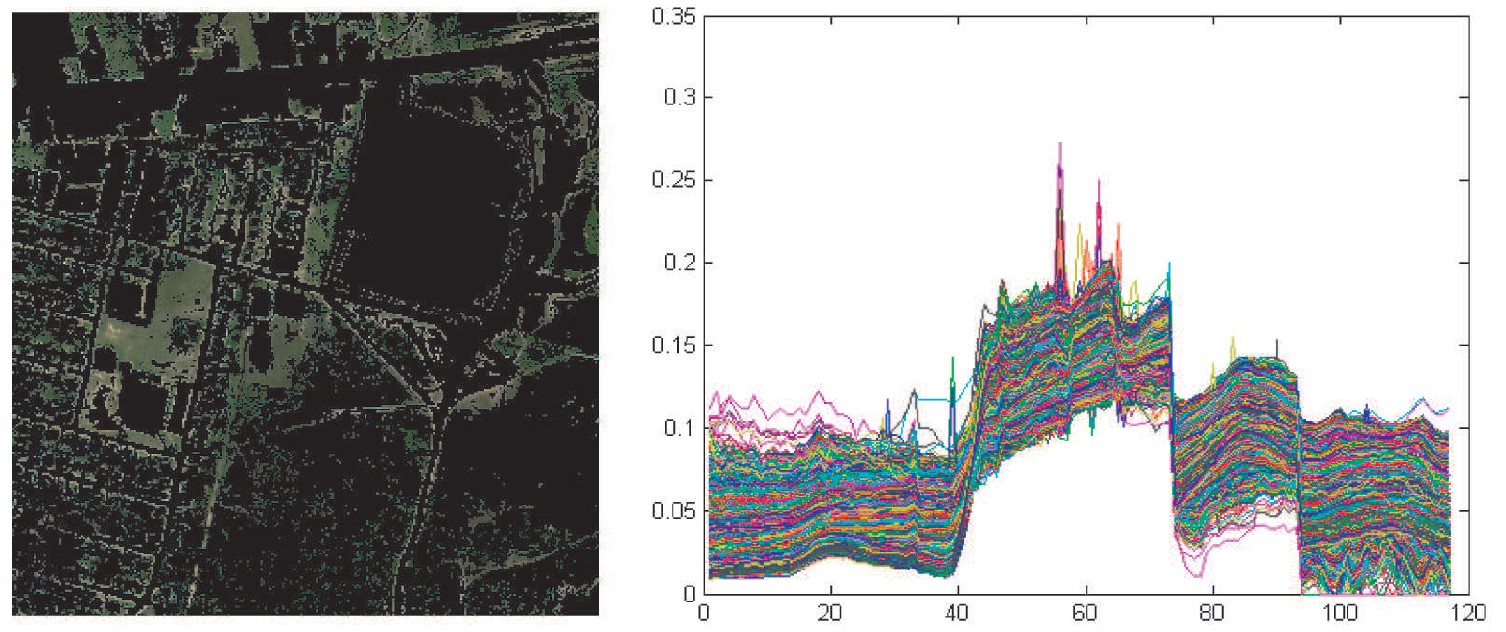}
   \end{tabular}
   \end{center}
\vspace{-10pt}
   \caption{ \label{Fig:5mowedgrass_gradientflow}
Gradient Flow clusters containing pixels from the Mowed Grass Class.}
\end{figure}

Figures~\ref{Fig:5mowedgrass_kmeans} and~\ref{Fig:5mowedgrass_gradientflow} show the clusters containing pixels from the Mowed Grass Class.
Both algorithms perform reasonably.
\vspace{-10pt}

\newpage
\subsection{Trees}
\begin{figure}[h]
   \begin{center}
   \begin{tabular}{c}
   \includegraphics[width=3.7in]{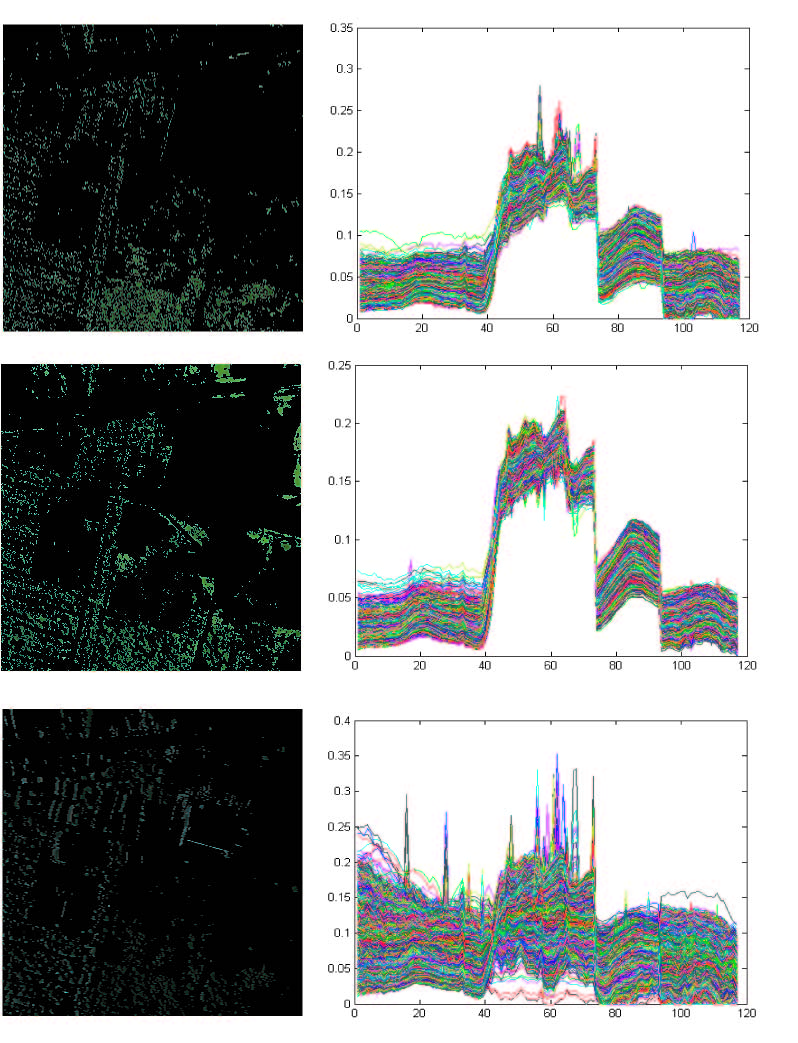}
   \end{tabular}
   \end{center}
\vspace{-10pt}
   \caption{ \label{Fig:6trees_kmeans}
$k$-means cluster containing pixels from the Tree Class.}
\end{figure}

\begin{figure}[h]
   \begin{center}
   \begin{tabular}{c}
   \includegraphics[width=3.7in]{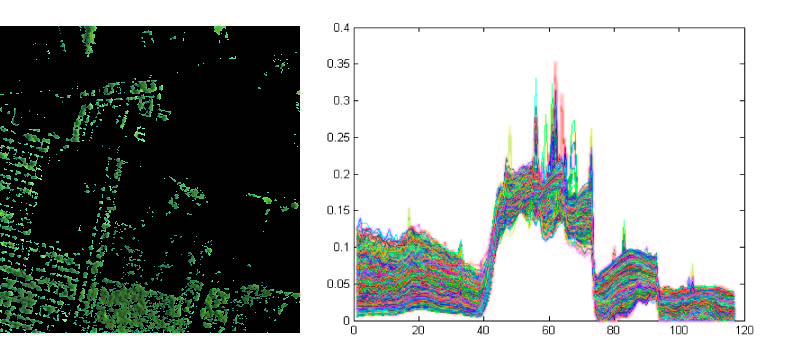}
   \end{tabular}
   \end{center}
\vspace{-10pt}
   \caption{ \label{Fig:6trees_gradientflow}
Gradient Flow clusters containing pixels from the Tree Class.}
\end{figure}

Figures~\ref{Fig:6trees_kmeans} and~\ref{Fig:6trees_gradientflow} show the clusters containing pixels from the Tree Class.  Both algorithms
perform reasonably, although the gradient flow cluster has better class preservation.
\vspace{-10pt}

\newpage
\subsection{Road 1 (Highway)}
\begin{figure}[h]
   \begin{center}
   \begin{tabular}{c}
   \includegraphics[width=3.7in]{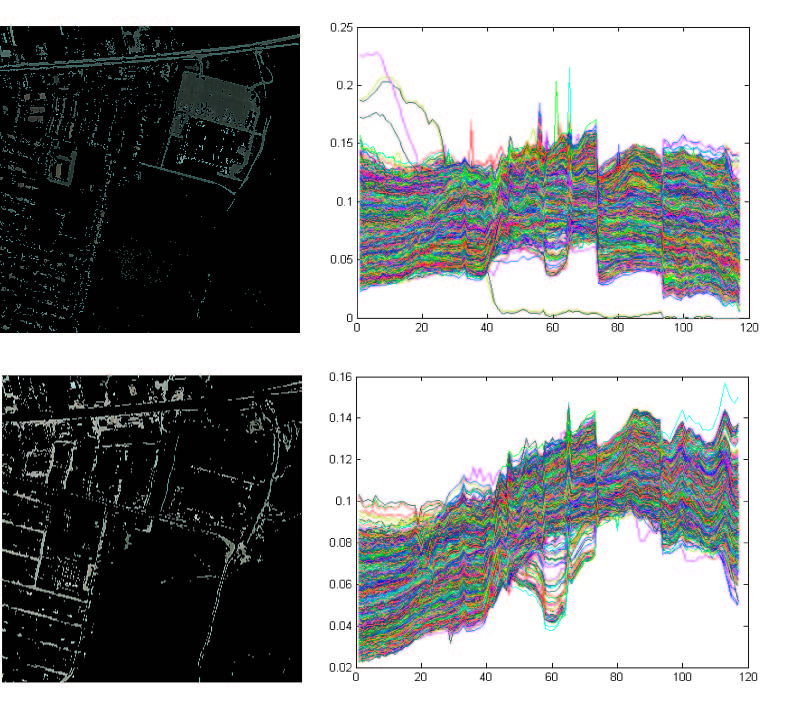}
   \end{tabular}
   \end{center}
\vspace{-10pt}
   \caption{ \label{Fig:7road1_kmeans}
$k$-means cluster containing pixels from the Road 1 (Highway) Class.}
\end{figure}

\begin{figure}[h]
   \begin{center}
   \begin{tabular}{c}
   \includegraphics[width=3.7in]{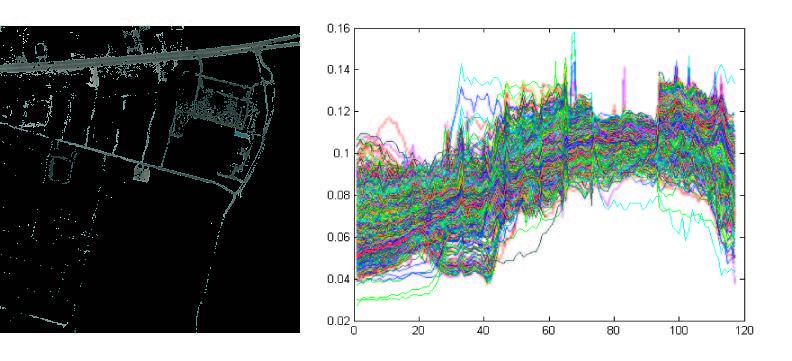}
   \end{tabular}
   \end{center}
\vspace{-10pt}
   \caption{ \label{Fig:7road1_gradientflow}
Gradient Flow clusters containing pixels from the Road 1 (Highway) Class.}
\end{figure}

Figures~\ref{Fig:7road1_kmeans} and~\ref{Fig:7road1_gradientflow} show the clusters containing pixels from the Road 1 (Highway) Class.  The
$k-$means algorithm has more confusion with road surfaces and roofs.
\vspace{-10pt}

\newpage
\subsection{Road 2 (Side Streets)}
\begin{figure}[h]
   \begin{center}
   \begin{tabular}{c}
   \includegraphics[width=3.7in]{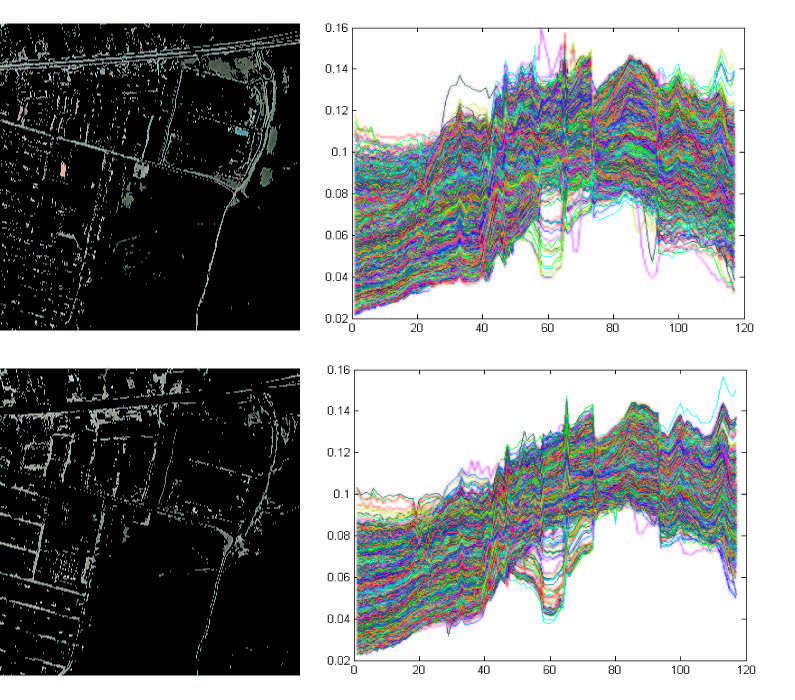}
   \end{tabular}
   \end{center}
\vspace{-10pt}
   \caption{ \label{Fig:8road2_kmeans}
$k$-means cluster containing pixels from the Road 2 (Side Streets) Class.}
\end{figure}

\begin{figure}[h]
   \begin{center}
   \begin{tabular}{c}
   \includegraphics[width=3.7in]{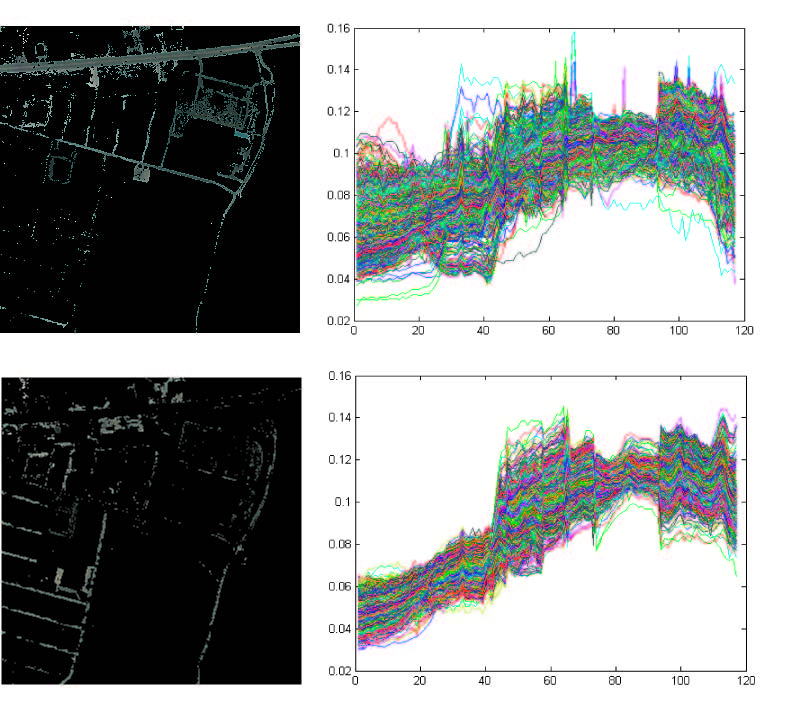}
   \end{tabular}
   \end{center}
\vspace{-10pt}
   \caption{ \label{Fig:8road2_gradientflow}
Gradient Flow clusters containing pixels from the Road 2 (Side Streets) Class.}
\end{figure}

Figures~\ref{Fig:8road2_kmeans} and~\ref{Fig:8road2_gradientflow} show the clusters containing pixels from the Road 2 (Side Streets) Class.
The $k-$means algorithm has more confusion with road surfaces and roofs.
\vspace{-10pt}

\newpage
\subsection{Parking Lot}
\begin{figure}[h]
   \begin{center}
   \begin{tabular}{c}
   \includegraphics[width=3.7in]{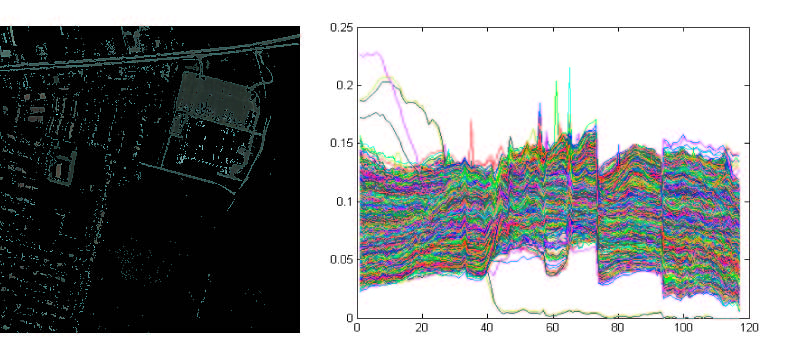}
   \end{tabular}
   \end{center}
\vspace{-10pt}
   \caption{ \label{Fig:9parkinglot_kmeans}
$k$-means cluster containing pixels from the Parking Lot Class.}
\end{figure}

\begin{figure}[h]
   \begin{center}
   \begin{tabular}{c}
   \includegraphics[width=3.7in]{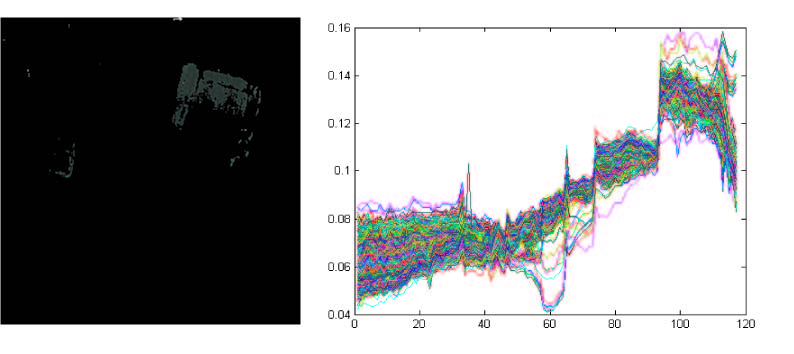}
   \end{tabular}
   \end{center}
\vspace{-10pt}
   \caption{ \label{Fig:9parkinglot_gradientflow}
Gradient Flow clusters containing pixels from the Parking Lot Class.}
\end{figure}

Figures~\ref{Fig:9parkinglot_kmeans} and~\ref{Fig:9parkinglot_gradientflow} show the clusters containing pixels from the Parking Lot Class.
Both algorithms preserve that parking lot class but the $k-$means algorithm includes a variety of road surfaces and roofs in this class.
The spectra for the gradient flow algorithm cluster appears particularly homogeneous.
\vspace{-10pt}

\newpage
\subsection{Soil}
\begin{figure}[h]
   \begin{center}
   \begin{tabular}{c}
   \includegraphics[width=3.7in]{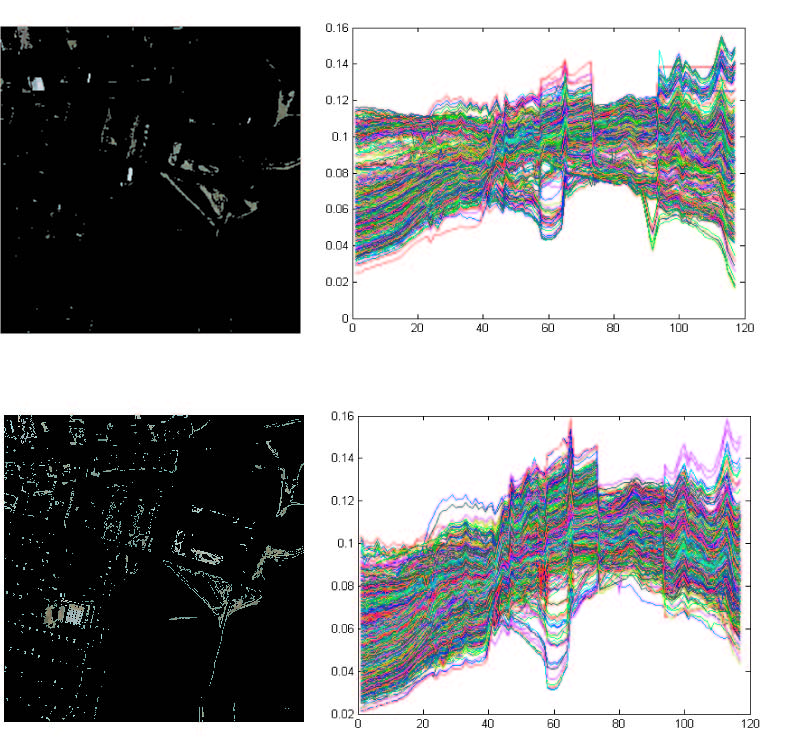}
   \end{tabular}
   \end{center}
\vspace{-10pt}
   \caption{ \label{Fig:10soil_kmeans}
$k$-means cluster containing pixels from the Soil Class.}
\end{figure}

\begin{figure}[h]
   \begin{center}
   \begin{tabular}{c}
   \includegraphics[width=3.7in]{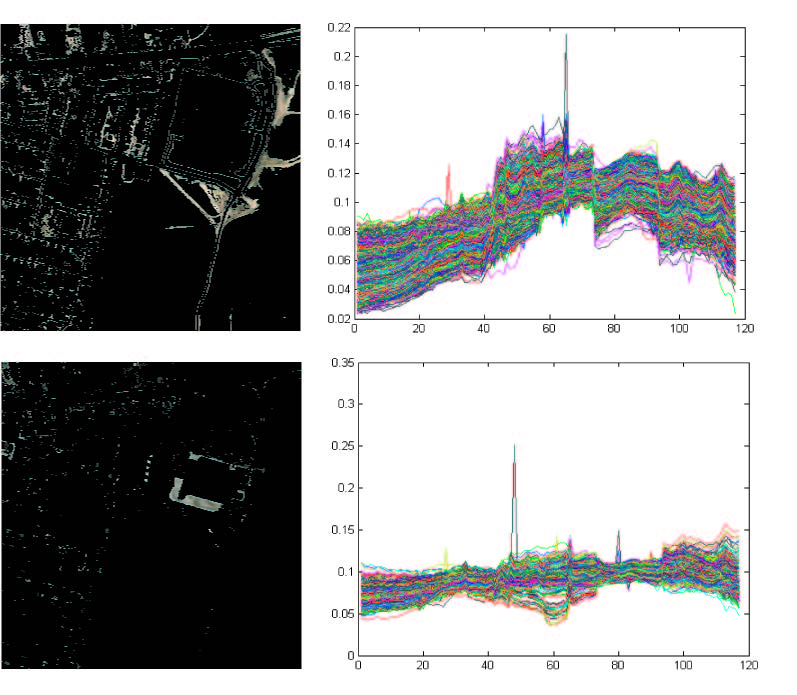}
   \end{tabular}
   \end{center}
\vspace{-10pt}
   \caption{ \label{Fig:10soil_gradientflow}
Gradient Flow clusters containing pixels from the Soil Class.}
\end{figure}

Figures~\ref{Fig:10soil_kmeans} and~\ref{Fig:10soil_gradientflow} show the clusters containing pixels from the Soil Class.  The spectra
suggest that the gradient flow algorithm has less mixing of spectra.
\vspace{-10pt}
\newpage



\bibliography{report}   
\bibliographystyle{spiejour}   

\end{document}